\documentclass[letterpaper]{article} 
\usepackage{aaai2026}  
\usepackage{times}  
\usepackage{helvet}  
\usepackage{courier}  
\usepackage[hyphens]{url}  
\usepackage{graphicx} 
\usepackage{natbib}  
\usepackage{caption} 
\usepackage{algorithm}
\usepackage{algorithmic}
\usepackage{amsmath}
\usepackage{arydshln}
\usepackage{tcolorbox}
\usepackage{booktabs}
\usepackage{array}
\usepackage{tcolorbox}
\usepackage{multirow}
\usepackage{amssymb}
\usepackage[utf8]{inputenc}
\usepackage{newfloat}
\usepackage{listings}
\DeclareCaptionStyle{ruled}{labelfont=normalfont,labelsep=colon,strut=off} 
\floatstyle{ruled}
\newfloat{listing}{tb}{lst}{}
\floatname{listing}{Listing}
\title{Look Closer! An Adversarial Parametric Editing Framework for Hallucination Mitigation in VLMs}
\author{
        Jiayu Hu\textsuperscript{\rm 1},
    Beibei Li\textsuperscript{\rm 1}\thanks{Corresponding author.},
    Jiangwei Xia\textsuperscript{\rm 1},
    Yanjun Qin\textsuperscript{\rm 2}, 
    Bing Ji\textsuperscript{\rm 1}, 
    Zhongshi He\textsuperscript{\rm 1}
}
\affiliations{
    \textsuperscript{\rm 1}College of Computer Science, Chongqing University\\
    \textsuperscript{\rm 2}School of Computer Science and Technology, Research Center for Multimodal Information Perception and Intelligent Processing, Xinjiang University\\
    $\{$hujiayu, libeibeics, zshe$\}$@cqu.edu.cn, xiajiangwei77@gmail.com, qinyanjun@xju.edu.cn
}

\usepackage{bibentry}

\begin{document}

\maketitle
\begin{abstract}
While Vision-Language Models (VLMs) have garnered increasing attention in the AI community due to their promising practical applications, they exhibit persistent hallucination issues, generating outputs misaligned with visual inputs. Recent studies attribute these hallucinations to VLMs' over-reliance on linguistic priors and insufficient visual feature integration, proposing heuristic decoding calibration strategies to mitigate them. However, the non-trainable nature of these strategies inherently limits their optimization potential. To this end, we propose an adversarial parametric editing framework for Hallucination mitigation in VLMs, which follows an \textbf{A}ctivate-\textbf{L}ocate-\textbf{E}dit \textbf{A}dversarially paradigm. Specifically, we first construct an activation dataset that comprises grounded responses (positive samples attentively anchored in visual features) and hallucinatory responses (negative samples reflecting LLM prior bias and internal knowledge artifacts). Next, we identify critical hallucination-prone parameter clusters by analyzing differential hidden states of response pairs. Then, these clusters are fine-tuned using prompts injected with adversarial tuned prefixes that are optimized to maximize visual neglect, thereby forcing the model to prioritize visual evidence over inherent parametric biases. Evaluations on both generative and discriminative VLM tasks demonstrate the significant effectiveness of ALEAHallu in alleviating hallucinations. 
\end{abstract}
\begin{links}
    \link{Code}{https://github.com/hujiayu1223/ALEAHallu}
\end{links}
\section{Introduction}
\begin{figure}[ht]
    \centering
    \includegraphics[width=0.445\textwidth]{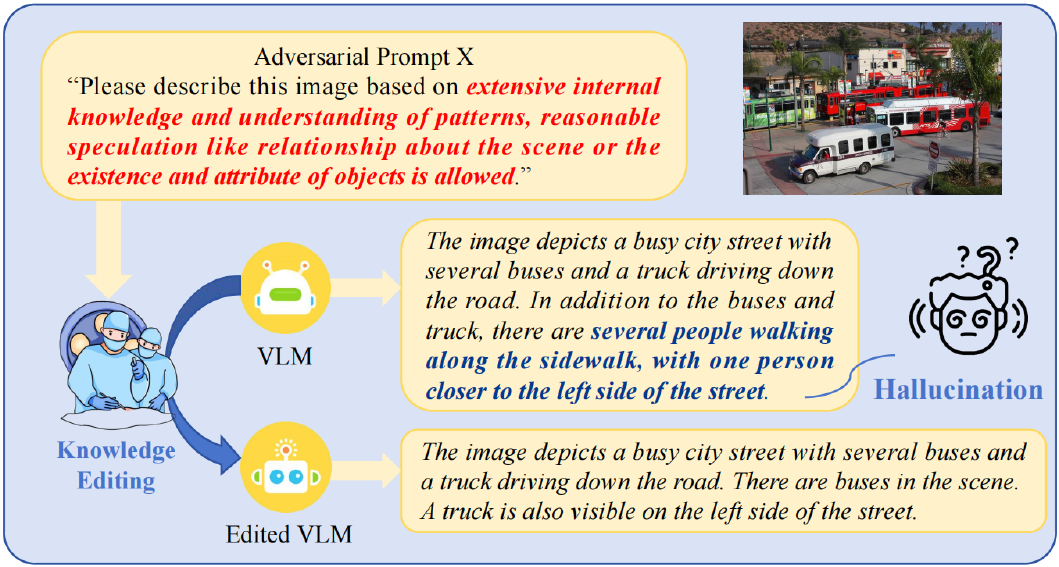} 
    \caption{An example of ALEAHallu.}
    \label{fig:illustration_s}
\end{figure}
In recent years, Vision-Language Models (VLMs) such as LLaVA \citep{liu2023visual}, MiniGPT-4 \citep{zhu2023minigpt}, and BLIP-2 \citep{li2023blip} have demonstrated remarkable capabilities across multimodal tasks including image-text generation, cross-modal retrieval, and visual question answering. Despite their impressive perceptual and generative capabilities, VLMs remain prone to hallucination issues \citep{liu2024survey}, where the model generates content inconsistent with the actual visual input on objects, attributes, and relationships, or even entirely unrelated. This issue greatly limits their reliability and performance in real-world applications.

Hallucinations in VLMs stem from multiple factors  \citep{liu2024survey}, including low-quality training data  \citep{liu2023mitigating}, imperfect cross-modality alignment \citep{zhao2023beyond}, limitations in visual encoder architecture \citep{zhai2023halle}, and biases inherent in Large Language Models (LLMs) \citep{ji2023survey}. A critical contributor is the insufficient integration of visual features during generation \citep{wang2024vigc}, coupled with an over-reliance on linguistic priors derived from text corpora \citep{liu2023mitigating}. This imbalance causes VLMs to prioritize linguistic fluency over factual accuracy, particularly when processing ambiguous visual inputs or out-of-distribution samples \citep{lee2023volcano}. In such cases, models often generate plausible yet factually inconsistent content that aligns with statistical language patterns rather than visual evidence \citep{wang2023evaluation}, leading to hallucinations.

Recent work has explored decoding strategies that modulate generation dynamics to recalibrate VLMs' attention toward visual context while reducing dependence on LLM priors \citep{leng2024mitigating,huang2024opera}. While they demonstrate effectiveness, their heuristic, non-trainable nature, relying on manual decoding rules rather than learnable parameters, often leads to suboptimal performance and increased inference overhead, highlighting the need for optimizable solutions. Conversely, knowledge editing, which is a parameter-efficient technique that surgically modifies a pre-trained model’s factual or behavioral knowledge at inference time without full retraining but by updating only a sparse set of its weights or activations, offers an efficient alternative. While knowledge editing has been applied to different LLM tasks  \cite{DINM, TruthX, TruthForest}, their potential for VLM hallucination mitigation remains largely unexplored. A very recent work  \cite{nullu} employ knowledge editing in hallucination mitigation but relies on SVD decomposition of model parameters, suffering from high computation cost.

To this end, we propose an simple yet efficient \emph{\textbf{A}ctivate--\textbf{L}ocate--\textbf{E}dit \textbf{A}dversarially} framework \textbf{ALEAHallu} that adversarially edits only those parameters responsible for vision hallucination. Specifically, we begin by curating an activation dataset of paired question-answer samples for each image: a \emph{positive} response that is strictly grounded in the visual evidence, and a \emph{negative} response that instead relies on parametric priors while ignoring the image.  Then, processing these pairs in parallel, we compare the resulting hidden-state activations and statistically isolate the sparse parameter clusters whose activity is most correlated with hallucinatory outputs. Finally, we formulate an adversarial objective that forces the model to generate faithful answers under prompts with an adversarial prefix, encouraging the model to prioritize visual cues and thereby reduce hallucinations under standard prompts. Particularly, instead of hand-crafting such an adversarial prefix, we treat the prefix itself as a learnable matrix: keeping the model parameters fixed, we optimize the prefix tokens to maximize the probability of hallucinated responses. 

Our contributions can be summarized as follows:
\begin{itemize}
\item \textbf{Dataset.} We construct a dataset specifically designed to localize hallucination-prone parameters, empowering systematic research into VLM hallucination mechanisms.
\item \textbf{Methodology.} We propose an efficient adversarial parametric editing framework ALEAHallu for  hallucination mitigation in VLMs, which follows an activate-locate-edit adversarially paradigm and edits only critical parameter clusters. 
    \item \textbf{Experiments.} Extensive experiments across generative and discriminative tasks confirm that ALEAHallu significantly reduces hallucinations and robustly refocuses model attention on visual evidence.
\end{itemize}
\section{Related Work}
\subsection{Visual-Language Models}

To extend the capabilities of LLMs to vision-language tasks, pre-trained visual feature alignment and visual instruction fine-tuning are commonly used to help LLMs to comprehend the format of instruction input and generate diverse content in a more comprehensive way by integrating information from both text and images.  Through these training process, more and more large visual language models (VLMs) emerged, including the series of CLIP  \cite{liu2024survey} and BLIP  \cite{li2023blip} well aligns the text features and image features and LLaVA  \cite{liu2023visual}, LLaVA-NeXT\cite{liu2024llavanext} and MiniGPT-4  \cite{zhu2023minigpt} could even allow users to interact with these intelligence with images and texts as prompts. Despite above advancements, specific challenges persist, especially the issue of object hallucination  \cite{gunjal2024detecting, li2023evaluating, lovenia2023negative} being a prominent concern that affects the reliability and applicability of VLMs across domains. Consequently, we propose a novel method to mitigate hallucinations in vision-language models in our paper.

\subsection{Hallucinations in VLMs}
Hallucination in vision-language models (VLMs) appears as a discrepancy between generated text and image facts. Its origins span limited vision encoders, cross-modal misalignment, data bias, annotation irrelevance, and intrinsic LLM hallucinations~ \cite{hu2023ciem,liu2023mitigating,liu2024investigatingmitigatingobjecthallucinations}. To alleviate this problem, recent research can be summarized into three lines of thought: at the training stage, by constructing positive and negative samples or introducing additional perceptual modalities to enhance the fine-grained nature and balance of visual instructions  \cite{liu2023mitigating, jain2024vcoder}, at the decoding stage, by leveraging contrasts among image–text~ \cite{zhu2024ibd}, inter-layer~ \cite{chuang2023dola}, or original-vs.-perturbed~ \cite{leng2024mitigating} visual distributions to suppress the model’s over-reliance on unimodal priors; other work attempts post-hoc correction with RLHF~ \cite{sun2023aligning} or modified beam search~ \cite{huang2024opera}. A very recent work~ \cite{DINM} eliminates VLM hallucinations via knowledge editing, yet they require SVD decomposition and incur heavy computational overhead.  In a word, these methods either computationally expensive or rely on heuristic principles, making them impractical in real-world applications and prone to suboptimal outcomes.  

\subsection{Knowledge Editing}
Knowledge editing techniques, which aim to efficiently adjust a model's behavior in specific domains while preserving overall performance, have garnered significant recent attention  \cite{zhang2024comprehensive}. Current methods for editing large models are primarily categorized into three types  \cite{zhang2024comprehensive}: Resorting to External Knowledge, Merging Knowledge into the Model, and Editing Intrinsic Knowledge. The first approach leverages prompting techniques, encouraging the model to reason using external knowledge provided within the input to generate improved responses  \cite{zheng2023can}. Methods focusing on merging knowledge internally integrate representations of new knowledge directly into the model's parameters; acknowledging that feed-forward networks (FFNs) store significant knowledge  \cite{geva2020transformer, geva2022transformer, chen2024journey}, several techniques specifically target modifications within these modules  \cite{dong2022calibrating, huang2023transformer}, while others directly replace specific internal representations  \cite{hernandez2023inspecting}. Editing Intrinsic Knowledge typically employs either Meta-Learning, utilizing an auxiliary hypernetwork to learn parameter adjustments based on new facts  \cite{de2021editing, hase2023methods}, or the Locate-and-Edit paradigm, which first identifies precise model locations storing specific knowledge before making targeted modifications  \cite{meng2022locating, meng2022mass}. Despite extensive research on knowledge editing for Large Language Models (LLMs), its application remains largely unexplored in Vision-Language Models (VLMs), particularly for addressing hallucination mitigation.

\begin{figure*}
	\centering
	\includegraphics[width=1\linewidth]{"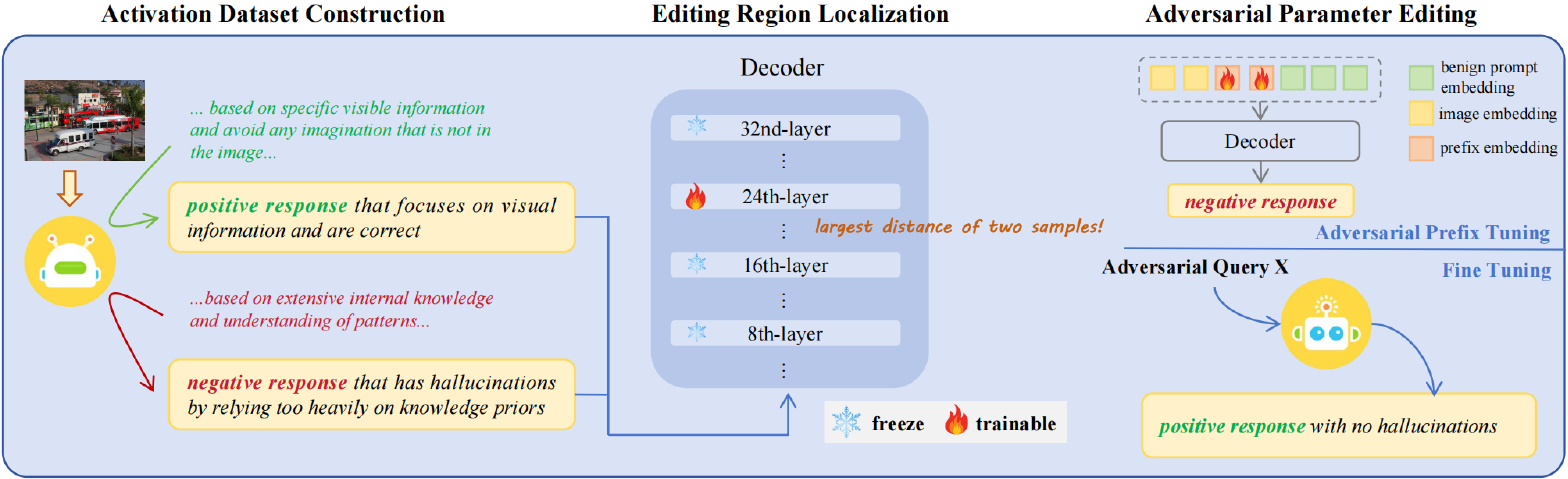"}
    \caption{ALEAHallu consists of three main components: activation dataset construction, editing region localization and adversarial parameter editing.}
    \label{fig:illustration}
\end{figure*}

\section{Methodology}
As illustrated in Figure \ref{fig:illustration}, ALEAHallu comprises three core components: activation dataset construction, editing region localization, and adversarial parameter editing. Specifically, given a query ( i.e., text prompt) and an image, we generate positive and negtive response pairs: a positive response that is visually grounded and hallucination-free, and a negative response that is over-reliance on LLM priors and containing hallucinations. These response pairs constitute the activation dataset and are utilized to pinpoint the critical hallucination-prone parameters that require editing. Subsequently, feeding the response pairs into the VLMs, we localize hallucination-prone parameters by analyzing the representational discrepancies between positive and negative samples in the latent space. Furthermore, we learn adversarial prompt prefix designed to overlook visual features and then edit  hallucination-prone  parameters to response correctly with adversarial prompt. 

\subsection{Activation Dataset Construction}
\paragraph{Prompt design.} To activate and localize hallucination-prone regions, we generate paired descriptions for the same image: one hallucination-free description that carefully attends to visual features, and one hallucinatory description containing erroneous objects, attributes, and relationships due to the neglect of visual information. Specifically, for each image $v$, we design a visually grounded prompt $x^+$ as shown in Prompt 1, focusing on visual features and eliciting a less hallucinated response $y^+$, forming the positive sample $[v, x^+, y^+]$. Conversely, for negative samples, we use an adversarial prompt $x^-$ as shown in Prompt 2, which focuses more on knowledge priors and is designed to trigger hallucinations, and concatenate it with the corresponding hallucinated response $y^-$ to form $[v, x^-, y^-]$. The detailed prompts are as follows.
\paragraph{GPT-4o assisted evaluation.}
To ensure that the positive sample does not contain hallucinations, we need to help evaluate the level of hallucination in the response. In this paper, we use GPT-4o as our evaluation tool. The detailed prompt is as Prompt 3.
\begin{tcolorbox}[title=\textbf{Prompt 1}: Focus on visual feature , colback=white, colframe=gray, colbacktitle=white!80!gray, coltitle=black,label = box1]
     Please describe the given image based on specific visible information and avoid any imagination that is not in the image.
\end{tcolorbox}
\begin{tcolorbox}[title=\textbf{Prompt 2}: Focus on language priors , colback=white, colframe=gray, colbacktitle=white!80!gray, coltitle=black]
     Please describe the given image based on extensive internal knowledge and understanding of patterns, reasonable speculation like relationship about the scene or the existence and attribute of objects is allowed.
\end{tcolorbox}


\begin{tcolorbox}[title= \textbf{Prompt 3}: GPT-4o evaluation prompt , colback=white, colframe=gray, colbacktitle=white!80!gray, coltitle=black]
You are required to score the hallucination degree of description of a given image. Please score 0$\sim$2 based on the principle of scoring.\\
\\
\textbf{Principle of scoring}:\\
\qquad $\bullet$ the description is \textit{precise and everything in the description is all visible in the image}: 0.\\
\qquad$\bullet$ there are \textcolor{blue}{\textit{wrong relationship and attribute of objects}} but the \textcolor{blue}{\textit{objects in description are all in the image}}: 1.\\
\qquad$\bullet$ there are \textcolor{red}{\textit{objects that do not exist in image but appear in the given description}}: 2.\\
\textbf{Please output the scores for the description}: $[$the generated image caption$]$.\\
Do not generate other sentence after the score.\\
\\
\textbf{Output format}:\\
Degree: $<$Scores of the description$>$
\end{tcolorbox}
 

 Samples with a score of 0 are hallucination-free and treated as positive samples. Samples scored 1 or 2 exhibit varying degrees of hallucinations and are considered negative samples. For an image, if the descriptions generated based on Prompt 1 and Prompt 2 correspond to hallucination-free and hallucinated outputs respectively, they form a positive-negative sample pair. In practice, we utilized Prompt 1 and Prompt 2 to generate descriptions for 4,000 images selected from MSCOCO. After verification by GPT-4o, 2,091 valid positive-negative sample pairs were obtained.

\subsection{Editing Region Localization} 

We adopt a simple strategy that leverages the representation of an entire response to localize the editing region. As we know, a Transformer consists of multiple decoder layers. During decoding, the input query prompt and the corresponding image are first encoded into embeddings and then forwarded through the layers. Each layer comprises a multi-head self-attention module and multi-layer perceptrons. 



To identify hallucination-prone regions, we feed the positive–negative response pairs constructed in the previous section into the VLM and compute the layer-wise Euclidean distance between their hidden representations. The layer exhibiting the largest distance is regarded as the hallucination-prone layer, denoted $\ell_{\text{hallu}}$:  
\begin{equation}
\ell_{\text {hallu }}=\underset{\ell \in\{1, \ldots, L\}}{\operatorname{argmax}}\left\|\mathbf{h}_{\ell}^{ (+)}-\mathbf{h}_{\ell}^{ (-)}\right\|_2,
\end{equation}
where $\mathbf{h}^{ (+)}_{\ell},\mathbf{h}^{ (-)}_{\ell}$ are the hidden states of the positive and negative responses at layer $\ell$, respectively.

Typically, the MLP in each decoder layer is a two-layer feed-forward network. Following prior work  \cite{DINM}, we treat the weight matrix of the second layer within the MLP in $\ell_{\text {hallu }}$-th layer as the hallucination-prone parameters, since this layer plays a pivotal role in knowledge dissemination during forward propagation and serves as the target of our subsequent editing.

\subsection{Adversarial Parameter Editing}

To edit the identified hallucination-prone region, we formulate an optimization objective that trains the model to remain robust, i.e., to generate hallucination-free content, even under adversarial input conditions. Such adversarial inputs include prompts that either encourage the model to ignore visual information or distract the model from attending to the correct visual cues. Our insight is that, \textit{to produce visually consistent outputs under such perturbations, the model is forced to attend more accurately to the visual evidence, thereby mitigating hallucinations under benign prompts, which are much easier than adversarial prompts}. Intuitively, if a person can answer questions correctly under strong distractions, she/he will certainly perform even better in distraction-free settings.

Let the benign prompt be $x$,  for example, in image-captioning, $x$ could be “\textit{Please describe the following image}”. We construct an adversarial prompt prefix, denoted $q$, e.g., “\textit{Please rely more on the LLM prior rather than the image content}”. Concatenating $q$ and $x$ yields the adversarial prompt $[q;x]$, which is supplied to the VLM together with the image $v$. We minimize the negative log-likelihood of the ground-truth non-hallucinated response $y^+$ under this adversarial prompt:  
\begin{equation}\label{eq:le}
\min_{\mathbf{W}_{t}}\; \mathcal{L}_{e} = -\log P_{\mathbf{W}_{t}}\bigl (y^+\mid v,q,x\bigr),
\end{equation}
where $\mathbf{W}_{t}$ denotes the hallucination-prone parameters at training step $t$. This objective encourages the model to generate accurate responses despite misleading instructions, progressively reducing hallucinations under benign scenarios.

Meanwhile, to prevent knowledge editing from distorting the original output distribution, we incorporate a KL-divergence regularization term  as Equation \ref{eq:lc}, which constrains the edited parameters $\mathbf{W}_{t}$ so that the model retains its ability to respond normally to benign user queries and to produce fluent sentences. This ensures that localized parameter updates do not degrade general performance.
\begin{equation}\label{eq:lc}
\min_{\mathbf{W}_{t}}\;\mathcal{L}_c=\operatorname{KL}\left (P_{\mathbf{W}^t}\left (\cdot \mid\left[v, x\right]\right) \| P_{\mathbf{W}_0}\left (\cdot \mid\left[v, x \right]\right)\right),
\end{equation}
where $\mathbf{W}_{0}$ denotes the original model parameters.

\paragraph{Adversarial Prefix Tuning.} 
Manually crafting adversarial prompt prefixes is  heuristics, labor-intensive and expertise-dependent. We therefore employ prompt tuning to automatically optimize an adversarial prefix. Our core idea involves: keeping the VLM parameters frozen, prepending a learnable prefix to the input prompt, and optimizing this prefix to induce hallucinatory responses resulting from the model's neglect of visual information. An adversarial prefix optimized under this objective is supposed to guide the prompt to disregard visual cues.

We parameterize the prefix as a continuous matrix $\mathbf{E}_{x}\in\mathbb{R}^{r\times d}$, where $r$ is the number of prefix tokens and $d$ is the word-embedding dimension of the VLM. This matrix is concatenated with the image embedding $\mathbf{V}$ and the benign prompt embedding $\mathbf{E}_{p}$, forming the input $[\mathbf{V};\mathbf{E}_{q};\mathbf{E}_{x}]$ to the VLM. To train the prefix, we freeze all the parameters of VLM and maximize the likelihood of the hallucinated response $y^-$ from our dataset as follows:  
\begin{equation}\label{eq:pre-tuning}
\min_{\mathbf{E}_{q}}\; \mathcal{L}_{q} = -\log P_{\mathbf{W}_{0}}\bigl (y^-\mid [\mathbf{V},\mathbf{E}_{q},\mathbf{E}_{x}]\bigr).
\end{equation}

Finally, when the edited model operates under benign prompts (significantly easier than adversarial prompts), the model will pay more attention to visual tokens and the hallucinations are mitigated naturally.
\begin{table*}[t]
\centering
\small
\setlength{\tabcolsep}{3pt}
\begin{tabular}{c|cccc|cccc|cccc}
\toprule
\multirow{2}{*}{Methods} &
\multicolumn{4}{c|}{LLaVa-1.5}&
\multicolumn{4}{c|}{LLaVa-NeXT} &
\multicolumn{4}{c}{MiniGPT-4} \\
& 
$\text{CHAIR}_s \downarrow$ & $\text{CHAIR}_i \downarrow$ & Recall$\uparrow$ & Len &
$\text{CHAIR}_s \downarrow$ & $\text{CHAIR}_i \downarrow$ & Recall$\uparrow$ & Len &
$\text{CHAIR}_s \downarrow$ & $\text{CHAIR}_i \downarrow$ & Recall$\uparrow$ & Len\\
\midrule
Regular     & 56.4 & 17.2 & 69.5 & 106.2 & 40.4 & 12.2 & 60.2 & 175.9 & 56.4 & 17.2 & 69.5 & 106.2\\
BS          & \underline{50.6} & \underline{13.5} & \textbf{78.8} & 97.6 &
              35.2 & \underline{8.5} & \underline{62.8} &  176.7 &
              \textbf{34.8} & \textbf{10.0} & 61.1 & 80.6\\
OPERA       & 51.3 & 13.6 & 74.2 & 93.6 &
              35.2 & \underline{8.5} & \underline{62.8} &  176.7 &
              \textbf{34.8} & \textbf{10.0} & 61.1 & 80.6\\
VCD         & 57.4 & 16.1 & \underline{76.7} & 103.1 &
              \underline{34.6} & 9.6 & 60.7 & 175.6 &
              57.4 & 16.1 & \underline{76.7} & 103.1\\
Nullu       & 46.8 & 14.9 & 67.9 & 95.0 &
              35.5 & 10.0 & 61.3 & 173.5 &
              48.4 & 13.5 & 56.8 & 105.2\\
ALEAHallu   & \textbf{39.3} & \textbf{12.1} & 74.2 & 98.1 &
              \textbf{34.4} & \textbf{7.6} & \textbf{63.0} & 185.6 &
              \underline{40.2} & \underline{13.0} & \textbf{78.0} & 98.1\\
\bottomrule
\end{tabular}
\caption{Results on image caption task using different base models, the values in bold and underlined are the best and second best results in each row.}

\label{exp:chair}
\end{table*}

\subsection{Training and Inference}

Our training proceeds in two stages.  
\emph{Stage 1}: We perform prefix tuning by optimizing Equation \ref{eq:pre-tuning} to obtain the adversarial prefix $\mathbf{E}_{q}$.  
\emph{Stage 2}: We jointly optimize the parameter editing objective $\mathcal{L}_{e}$ together with the KL regularization term $\mathcal{L}_{c}$ as follows:  
\begin{equation}
\min_{\mathbf{W}_{t}}\mathcal{L} = \min_{\mathbf{W}_{t}} (\lambda\mathcal{L}_{e} + \mathcal{L}_{c}),
\end{equation}
where $\lambda$ is a scalar weight.

At inference time, we simply feed the image and the benign prompt into the VLM and generate output.Note that during the training phase we freeze every parameter except those in the hallucination-prone region, and our method is identical to the original model at inference and incurs zero additional decoding overhead. Hence, it is highly efficient.

\section{Experiment}

\subsection{Settings}

\subsubsection{Datasets and Metrics}
\paragraph{Image Caption Tasks.} We conduct experiments of image caption tasks on the MSCOCO dataset  \cite{lin2014microsoft} by querying the same prompt 'Please describe this image in detail', which contains over 300,000 images and 80 object categories with annotations. We use CHAIR\citep{rohrbach2018object} to to quantify the degree of object hallucination in image captioning by determining the proportion of objects mentioned in the generated description but absent from the ground-truth set. It provides two metrics: $\text{CHAIR}_s$ and $\text{CHAIR}_i$, to measure hallucinations at the sentence and image levels, respectively. We also evaluate the responses using Recall and description length. 


\paragraph{VQA Tasks.} We choose POPE evaluation, which designs binary questions about object presence in images. They all include three sampling settings: random, popular, and adversarial. We select the MSCOCO, A-OKVQA  \cite{schwenk2022okvqa}, and GQA  \cite{hudson2019gqa} for our POPE benchmark and adopt Accuracy, Precision, Recall, and F1 score as our evaluation metrics. The three sources involve 500 images from each dataset under each sampling setting, with six questions formulated per image. Since the negative samples can be easily constructed by flipping the binary "yes/no" answers from the corresponding positive samples in this setting, we select 2,000 pairs and use 80\% of them as training data, 20\% of them as test data. 

For overall performance assessment, we introduce The Multi-modal Large Language Model Evaluation (MME) benchmark  \cite{mme}, which contains ten perception-related and four cognition-related tasks, encompassing both object-level and attribute-level hallucinations. We select perception-related tasks in our experiments to test the edited models. 

\subsubsection{Compared Methods}

We compare ALEAHallu with five baselines: 1) \textbf{Direct Sampling}: The next token is directly sampled from the post-softmax distribution. 2) \textbf{Beam Search (short as BS)}   \cite{wu2016google}: It keeps the top-k most likely sequences at each step instead of just the best one and expands all current beams by possible next tokens and retains the top-k overall.  3) \textbf{OPERA}   \cite{huang2024opera}: It utilize a penalty term on the model logits during the beam-search decoding, along with a rollback strategy that retrospects the presence of summary tokens in the previously generated tokens, and re-allocate the token selection if necessary. 4) \textbf{VCD}   \cite{leng2024mitigating}: It calibrates the model’s outputs by contrasting output distributions derived from original and distorted visual inputs. 5) \textbf{Nullu}  \cite{nullu}: It identifies a subspace by extracting the hallucinated embeddings features and removing the truthful representations. Input features will be projected into the Null space of it by orthogonalizing the model weights.

\subsubsection{Implementation Details.}
We choose 500 positive and negative sample pairs to locate the editing regions and train our model on 8 NVIDIA GeForce RTX 3090 Ti GPUs, setting the batch size, epochs to 10, 5 respectively, each epoch runs about 1045 seconds. We adopt LLaVa-1.5, LLaVa-NeXT and MiniGPT-4 as base model in our experiments and set LLaVA-1.5 as default one. We set the learning rate and weight decay to $2 \times 10^{-5}$, max new tokens is set to 512, $\lambda$ is set to 0.1. For Beam Search and OPERA decoding, we set the number of beams to 2, VCD and Nullu are under their default configurations. The length of our learnable prefix is 5 tokens. 1,500 samples are used to optimize the prefix.

\subsection{ Experimental Results}

\paragraph{Results on Image Caption Tasks.} We provide the experimental results of image caption tasks on three models: LLaVa-1.5, LLaVa-NeXT and MiniGPT-4 as shown in Table \ref{exp:chair}. ALEAHallu maintains the lowest or second-lowest hallucination rates while achieving the highest recall among all methods. 

For the metrics $\text{CHAIR}_s$ and $\text{CHAIR}_i$ on LLaVa-1.5, the two knowledge editing methods, i.e., Nullu and ALEAHallu, outperform all other compared method based on decoding strategy, indicating that knowledge editing has advantages in reducing object hallucinations and our method could achieve the better performance. Regarding Recall, beam search achieves the highest score of 78.8 on LLaVa-1.5, suggesting that it can generate the most ground-truth objects from the images. Our method achieves a Recall of 74.2, slightly lower but still outperforms the other baselines. This may be attributed to the broader search space provided by beam search, allowing it to capture a wider range of objects across different generated sequences.

For LLaVa-NeXT, hallucinations decrease across all methods on this stronger model and our method still delivers the best performance——the lowest $\text{CHAIR}_s$, $\text{CHAIR}_i$ and the highest Recall. Meanwhile, the sequence lengths generated by ALEAHallu are comparable to those of other approaches, indicating that the decreasing in $\text{CHAIR}_s$ and $\text{CHAIR}_i$ are not simply due to longer descriptions, thereby ensuring a fair comparison. 

As for MiniGPT-4, although certain baselines (e.g., BS and OPERA) obtain slightly lower $\text{CHAIR}_i$ than ours, their recall drops substantially (e.g., 10 vs. 13 for $\text{CHAIR}_i$, but recall only 61.1 vs. 78 for ALEAHallu). We could also find that OPERA has limited ability in this model as well as MiniGPT-4 and obtain the same results as Beam Search's.


\begin{table}[t]
\centering
  
  \scalebox{0.9}{\small
\begin{tabular}{c|c|c|c|c}
    \toprule
        Datasets & Methods & Random & Popular & Adversarial \\ \midrule
        \multirow{5}{*}{MSCOCO} & ALEAHallu & \textbf{0.9072 } & \textbf{0.8952 } & 0.8016  \\ \cmidrule{2-5}
        & Regular & 0.8452 & 0.8288  & 0.8076\\ 
         & Beam Search & \underline{0.8752 } & \underline{0.8692 } & \textbf{0.8528 } \\
         & OPERA & 0.8744  & 0.8684& \underline{0.8520 } \\ 
         & VCD & 0.8652  & 0.8464  & 0.8252  \\
         
         ~&Nullu & 0.8499&0.8200&0.7900\\\midrule
         
        \multirow{5}{*}{A-OKVQA} & ALEAHallu & 0.8832  & \textbf{0.8724 } & \textbf{0.8020 } \\ \cmidrule{2-5}
        & Regular & 0.8516  & 0.8088  & 0.7388 \\ 
        ~ & Beam Search & 0.8852  & 0.8412 & 0.7812  \\ 
        ~ & OPERA & \underline{0.8868 } & 0.8420  & \underline{0.7856} \\ 
        ~ & VCD & 0.8696  & 0.8124  & 0.7488  \\ 
        ~&Nullu&\textbf{0.8957}&\underline{0.8600}&0.7800\\
         \bottomrule
    \end{tabular}
    }
    
    \caption{Results of Accuracy on VQA tasks, the values in bold and underlined are the best and second best results in each column.}
    
  \label{exp:pope}
\end{table}

\paragraph{Results on VQA Tasks.}
 According to the results in Table \ref{exp:pope}, ALEAHallu generally outperforms other approaches across different settings, particularly under the popular scenario, demonstrating a clear advantage in reducing object hallucinations. Our approach is also competitive in rest cases, highlighting its robustness across diverse query distributions. While ALEAHallu delivers the best overall performance, Beam Search and OPERA decoding also show their advantages in certain settings, indicating that decoding strategies still play a role in mitigating hallucinations to some extent.
\begin{figure}[t]
	\centering
	\includegraphics[width=0.8\linewidth]{"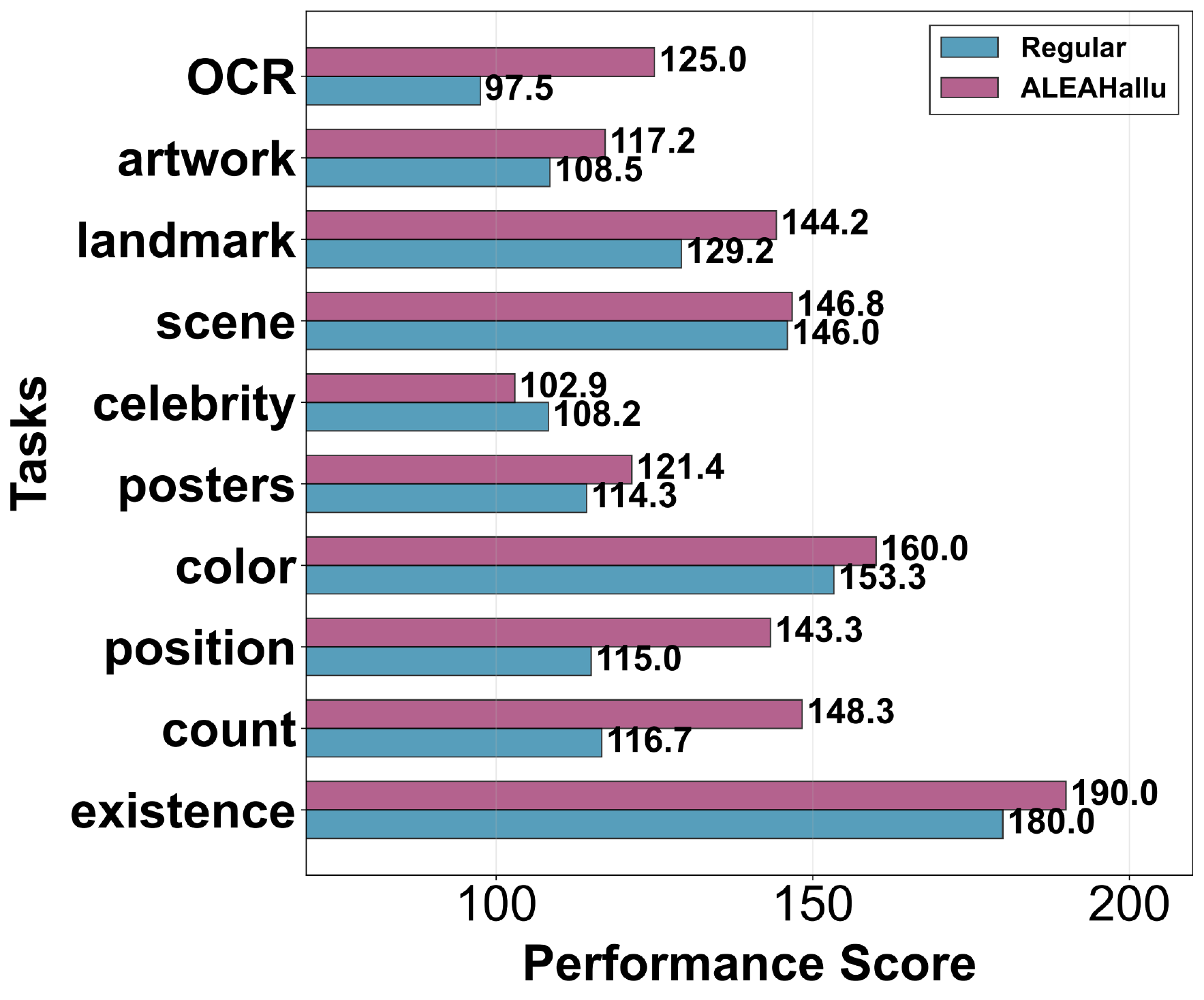"}
    \caption{Results on MME.}
    \label{exp:MME}
\end{figure}
\paragraph{Results on MME.}
 We present the results of ten perception-related tasks in Figure \ref{exp:MME} and use LLaVA-1.5 as a representative. Our methods achieves better performance on most tasks, especially for \textit{color}, \textit{position},\textit{count} and \textit{existence} task which evaluates the performance in reducing object-level and attribute-level hallucinations. The other tasks results also shows that our method leads to an enhancement of the general capability in perception-related tasks. 
 
These results suggest that ALEAHallu not only effectively aligns generated content with visual features but also generalizes well across different hallucination scenarios, validating its effectiveness in addressing the core challenges of hallucination in VLMs.

\begin{table}[t]
   
    \centering
    \small
    \setlength{\tabcolsep}{3.5pt} 
    \begin{tabular}{c|cccc}
    \toprule
        Methods  & $\text{CHAIR}_s \downarrow$ & $\text{CHAIR}_i \downarrow$ & Recall $\uparrow$& Len \\ 
        \midrule
        
         ALEAHallu & \textbf{39.3}  & \textbf{12.1} & \textbf{74.2} & \textbf{92.2} \\ \cmidrule{1-5}
        Regular & 56.4 & 17.2& 69.5&106.2 \\ 
        w/o Tune & 45.8  & 12.9 & 78.5 & 93.7 \\ 
        w/o Location & 55.4 &16.1 &78.4 &100.5 \\
        w/o Editing  & 52.0  & 15.3 & 79.3 & 97.0 \\ 
         w/o Constraint & 48.2  & 13.6 & 77.1 & 100.5 \\ 
        w/o Prefix Tuning & 48.4 & 13.0 & 78.0 & 98.1 \\
         \bottomrule
    \end{tabular} 
     \caption{Ablation study.}
     
    \label{exp:ablation}
\end{table}

\subsection{Ablation Study}
We assess the effectiveness of each component in ALEAHallu in ablation study. We evaluate the impact of parameter tuning, where only the prompt that focuses more on the visual feature is used without fine-tuning the model (w/o Tune). Additionally, we investigate the effect of locating the hallucination region (w/o Location) by randomly selecting one layer for editing instead of using our proposed identification method. To assess the contribution of the loss functions, experiments are conducted by individually removing the knowledge editing loss ($\mathcal{L}_e$), the knowledge constraint loss ($\mathcal{L}_c$) and prompt learning, denoted as w/o Editing, w/o Constraint and w/o Prefix Learning, respectively. The results are shown in Table \ref{exp:ablation}.

\paragraph{Editing and Constraint losses play important roles in ALEAHallu.}

Omitting parameter tuning results in significantly higher $\text{CHAIR}_s$ and $\text{CHAIR}_i$ scores, indicating increased object hallucination. This demonstrates the necessity of targeted fine-tuning. Additionally, removing either $\mathcal{L}_e$ or $\mathcal{L}_c$ results in reduced performance, they still have a better performance than regular method but worse than ALEAHallu, even though w/o Editing obtains higher Recall, that is because without the restraint of $\mathcal{L}_c$, it could not only generate more possibly existed object but also more hallucinated objects, highlighting the complementary roles of both losses in effectively guiding the training process.

\begin{table}[ht]

    \centering
    \scalebox{1}{\small
    \setlength{\tabcolsep}{1.3pt} 
    \begin{tabular}{c|c|c|c|c|c}
    \toprule
        Dataset& Model & Accuracy & Precision & Recall & F1 score \\ \midrule
        \multirow{3}{*}{MSCOCO}& $E_{\text{A-OKVQA}}$ & \textbf{0.9060} & \textbf{0.9042} & \textbf{0.9108} & \textbf{0.9074} \\ 
        & $E_{\text{GQA}}$ & 0.8948  & 0.8942  & 0.8263 & 0.8886 \\ 
        & Regular & 0.8452 & 0.8488 & 0.8316 & 0.8502 \\ \midrule
        \multirow{3}{*}{A-OKVQA}& $E_{\text{MSCOCO}}$ & 0.8732 & 0.8234 & \textbf{0.9568} & 0.8850 \\ 
        & $E_{\text{GQA}}$ & \textbf{0.9100} & \textbf{0.9022} & 0.9239  &\textbf{ 0.9128} \\ 
        & Regular & 0.8516 & 0.8256 & 0.8820 & 0.8701 \\ \midrule
        \multirow{3}{*}{GQA}& $E_{\text{MSCOCO}}$ &\textbf{ 0.8736 } & \textbf{0.8212 } & 0.9616 & \textbf{0.8858} \\ 
        & $E_{\text{A-OKVQA}}$ & 0.8516 & 0.7862 & \textbf{0.9742 } & 0.8701 \\ 
        & Regular & 0.8324 & 0.7900 & 0.8964 & 0.8632 \\ \bottomrule
    \end{tabular}
    }\caption{The generalization ability across datasets.}
    
    \label{exp:generalization_pope}
\end{table}

\paragraph{Locating the correct layers for editing is essential.}
Editing a randomly selected layer instead of our localization approach leads to a noticeable performance drop, confirming the importance of accurate hallucination region identification. Specifically, the higher layers which contains most semantic information are selected more in our experiments, randomly choose one layer in shallow, middle layers would not well guide the representation to a more image focus direction. 
\paragraph{Prefix Tuning helps enhance the capability of ALEAHallu.}
Using manually designed prompt rather than learnable prompt would reduce performance in generating non-hallucinatory responses, however, it could capture more existent objects information in images since its Recall is higher than ALEAHallu.

\subsection{Generalization Analysis}
We conduct cross-dataset experiments on the POPE benchmark to investigate whether edits based on one dataset (e.g., MSCOCO) can generalize to other datasets (e.g., GQA). Specifically, we train three edited models, i.e., $E_{\text{MSCOCO}}$, $E_{\text{A-OKVQA}}$, and $E_{\text{GQA}}$, each fine-tuned on hallucination-prone samples from the respective dataset. The results are shown in Table \ref{exp:generalization_pope}. It demonstrates that our editing model, when applied to one dataset, could also enhance the ability to mitigate hallucinations between datasets, which shows the superiority of ALEAHallu. 

\subsection{Visual Attention Analysis}
\begin{figure}
    \centering
    \includegraphics[width=0.93\linewidth]{"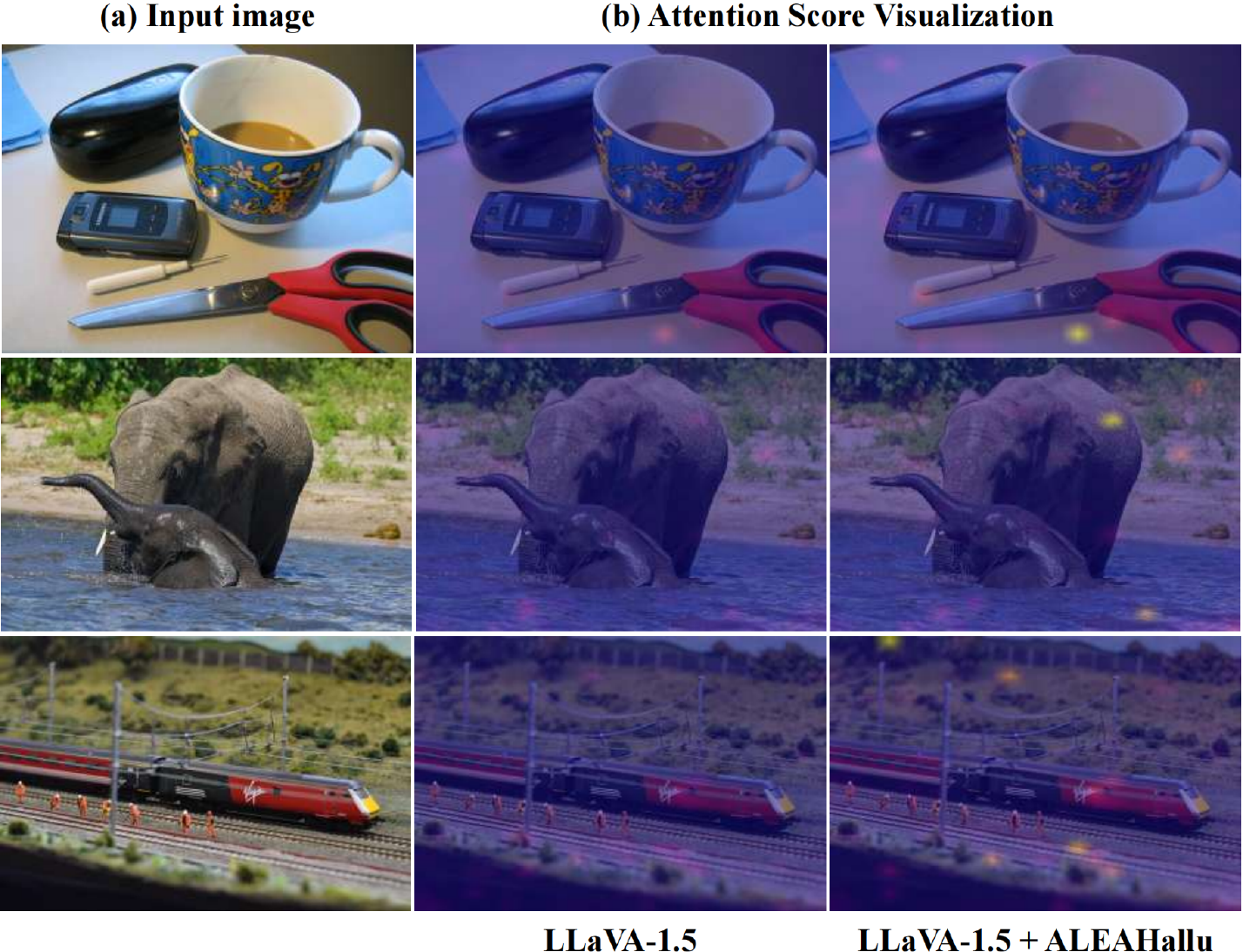"}
    \caption{Visualization of Visual Attention Scores.}
    \label{fig:visual_attention}
\end{figure}
This paper aims to enhance visual attention in VLM through knowledge editing, thereby mitigating hallucinated outputs. Experimental results on a test set of 500 samples show that the average proportion of attention allocated to the image by the VLM increased from 29.71\% to 33.72\%.

To provide a more intuitive understanding of the change in attention before and after knowledge editing, we visualize the attention scores between the last input token and each image token. In these visualizations, brighter regions indicate higher attention weights. As illustrated in Figure \ref{fig:visual_attention}, the high-attention regions become both more prominent and more widespread after editing, suggesting that the VLM pays significantly more attention to the image content following the knowledge edit.
\section{Conclusion}

In this paper, we address the persistent hallucination problem in VLMs and propose \textbf{ALEAHallu}, an adversarial parametric editing framework for hallucination mitigation. By following an \textit{activate-locate-edit adversarially paradigm}, ALEAHallu efficiently reduces hallucinations through edits to only a minimal parameter subset, while imposing no additional inference overhead. Extensive experiments demonstrate that ALEAHallu outperforms existing approaches while significantly enhancing VLMs' attentional focus on visual evidence during generation, thereby validating its effectiveness. For future work, we plan to integrate curriculum learning to further explore the performance ceiling of knowledge editing for hallucination suppression in VLMs.

\section{Broader Impacts}

Our work propose a hallucinations mitigating method for visual language models, which has a broader impact on enhancing trust in AI systems by reducing misinformation, making VLMs safer and more reliable in practical application. particularly in high-stakes domains like healthcare, education, and media. In education, hallucination mitigation may enhance learning tools by grounding visual-textual explanations in factual content. Nevertheless, over-mitigation of hallucinations could introduce trade-offs. In creative domains, suppressing imaginative synthesis may limit expressive potential. The societal impact of hallucination mitigation ultimately depends on balancing factual alignment with transparency, interpretability, and the preservation of human agency in evaluating AI outputs.

\section*{Acknowledgements}

This work was supported by the Chongqing Science and Technology Bureau (CSTB2022TTAD-KPX0180); “Tianchi Yingcai” Introduction Program;the National Natural Science Foundation of China Nos.62306164; Basic Research Project of the Autonomous Region's Universities' Basic Research Operating Funds (XJEDU2025J001); Open Research Fund Program of Beijing National Research Center for Information Science and Technology; Key Research and Development Project of the Autonomous Region(2024B03028).

\bibliography{aaai2026}

\end{document}